\documentclass[letterpaper, 10 pt, conference]{ieeeconf}  
\IEEEoverridecommandlockouts                              
\makeatletter
\let\NAT@parse\undefined
\makeatother

\usepackage[numbers,sort&compress]{natbib}

\usepackage{multicol}
\usepackage[bookmarks=true,urlcolor=blue,colorlinks=true]{hyperref}
\usepackage{graphics} 
\usepackage{epsfig} 
\usepackage{times} 
\usepackage{amsmath} 
\usepackage{amssymb}  
\usepackage{mathtools}
\usepackage{siunitx}
\usepackage{bm}
\usepackage{enumerate}

\title{\LARGE \bf
Feedback Linearization for Quadrotors with a Learned Acceleration Error Model
}

\author{Alexander Spitzer and Nathan Michael
  \thanks{Authors, \{spitzer,nmichael\}@cmu.edu, are with The Robotics Institute, Carnegie Mellon University,
          Pittsburgh, PA 15213, USA}
}

\begin{document}

\maketitle
\thispagestyle{empty}
\pagestyle{empty}

\begin{abstract}
This paper enhances the feedback linearization controller for multirotors with a learned acceleration error model and a thrust input delay mitigation model.
Feedback linearization controllers are theoretically appealing but their performance suffers on real systems, where the true system does not match the known system model.
We take a step in reducing these robustness issues by learning an acceleration error model, applying this model in the position controller, and further propagating it forward to the attitude controller.
We show how this approach improves performance over the standard feedback linearization controller in the presence of unmodeled dynamics and repeatable external disturbances in both simulation and hardware experiments.
We also show that our thrust control input delay model improves the step response on hardware systems.

\end{abstract}

\section{INTRODUCTION}

Quadrotors are extensively used in a variety of applications that require accurate control, such as inspection \cite{omari_visual_2014}, indoor and subterranean exploration \cite{tabib_autonomous_2020}, and cinematography \cite{bonatti_autonomous_2018}.
Inaccurate dynamical models impede the ability of control algorithms to generate accurate control inputs.
Furthermore, many practical applications use vision-based algorithms for state estimation \cite{tabib_autonomous_2020, bodie_omnidirectional_2019}.
These algorithms can be brittle to environmental conditions and can generate drifting or discontinuous state estimates.
Similarly, GPS-based state estimation can exhibit jumps when the signal quality is degraded, leading to jumps in tracking error.
Unanticipated tracking error can also arise from gusts of wind and other external disturbances.
Thus, the control algorithms used must be able to handle both poorly estimated dynamics and jumps in the state estimate.

The most common multirotor feedback controller discussed in the literature and employed in practice is one based on \textit{backstepping}, or a \textit{cascaded} controller, where a position control loop generates references for an attitude control loop.
Such cascaded controllers often make the assumption that the inner attitude control loop can track references much faster than the outer position control loop.
While approximately correct, this assumption introduces non-ideal characteristics in flight during aggressive or highly dynamic vehicle maneuvers, where the attitude tracking error is likely to be large.
In contrast, feedback linearization exactly linearizes the multirotor system,
leading to a linear error response.

A linear error response is advantageous for several reasons.
First linear controllers are easy to tune using well known traditional control techniques such as LQR and pole placement.
Second, a linear error response ensures exponential convergence to the reference state.
For example, for a quadrotor, a step input in position results in the vehicle taking the linear path connecting the start and end positions.
Further, any jumps in the state estimate, or disturbances causing large transient error, will be handled linearly.
The traditional cascaded quadrotor controllers do not have this property and their use may result in instability when the error is large.

\begin{figure}
  \begin{center}
    \includegraphics[width=0.48\textwidth]{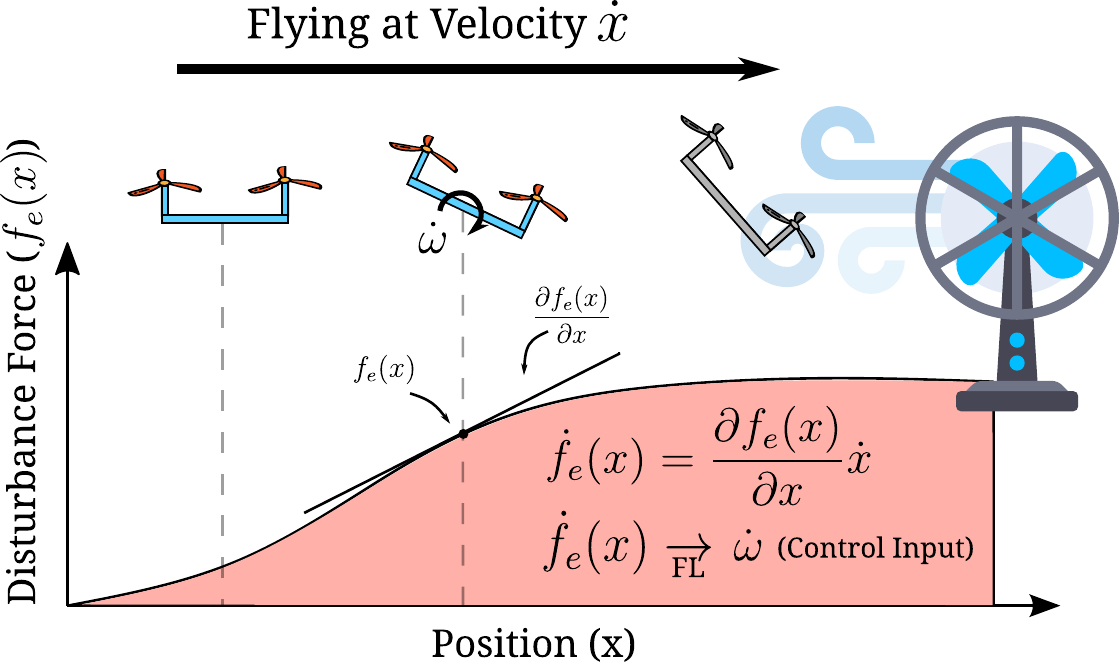}
  \end{center}
  \caption{
  An illustration of the proposed acceleration disturbance compensation method.
The vehicle is flying with velocity $\dot x$ head on into a strong wind field.
The acceleration disturbance from the wind field is modeled as a function $f_e(x)$ that depends on the vehicle position among other quantities.
Using the derivatives of the learned model, $\frac{\partial f_e(x)}{\partial x}$, and the vehicle velocity $\dot x$, the jerk disturbance $\dot f_e(x)$ can be computed.
The jerk disturbance, which predicts how the disturbance will change, is then used in a feedback linearizaton (FL) controller that computes the angular acceleration control input $\dot \omega$ to mitigate the effects of the wind.
  }
  \label{fig:pictoral}
\end{figure}

However, feedback linearization is known to be brittle when the dynamical model of the system is not well known.
In this work, we derive a feedback linearization controller for the multirotor and show how it can be used with a learned acceleration error model to improve performance in the presence of unmodeled or mismodeled dynamics and external disturbances.
We use incremental linear regression with a nonlinear feature space to rapidly learn a disturbance model over the state-space online.
This model and its derivatives are then used in the feedback linearization transformations to account for and anticipate changes in the disturbance, as shown in Fig.~\ref{fig:pictoral}.

In addition to correcting for unmodeled acceleration, we also correct for thrust control input delay.
Specifically, we show that control input delay, as is always present in real systems, can significantly impact the performance of the multirotor feedback linearization controller, and is often not considered in quadrotor control system design.
To mitigate this, we augment the multirotor model with a delay in the thrust control input, and investigate the performance of the resulting feedback linearization controller in both simulation and hardware.

The contributions of this paper are as follows:
\begin{enumerate}
  \item A \textbf{simple, Euler-angle free, derivation of the feedback linearization controller} for the multirotor with torque control inputs.
  \item A \textbf{control input delay mitigation model} for feedback linearization that improves performance in hardware experiments.
  \item A \textbf{learned acceleration model correction} that is used in the feedback linearization controller to fully compensate for the unmodeled dynamics and external disturbances. This is an extension of the work in \cite{spitzer_inverting_2019} for the feedback linearization controller.
\end{enumerate}
We experimentally analyze the effectiveness of the acceleration error compensation and input delay mitigation model in the feedback linearization controller on a real hardware quadrotor.

\section{Related Works}

Feedback linearization \cite{isidori_nonlinear_1995} has been applied to multirotors since at least \citet{mistler_exact_2001}, although the same technique has been used much earlier in the context of simplified helicopter models \cite{koo_output_1998}.
\citet{chang_global_2017} extends feedback linearization to preserve the positive thrust constraint using a modified dynamic extension and apply feedback linearization chartwise to avoid singularities.
\citet{lee_feedback_2009} compares feedback linearization to sliding mode control for a multirotor in the context of robustness, although a small-angle approximation is used.

Some existing works that present feedback linearization for multirotors
    do not present hardware results \cite{chang_global_2017, akhtar_path_2012},
    do not compare to cascaded approaches \cite{raffler_path_2013, achtelik_inversion_2013, holzapfel_novel_2011},
    or only use angular velocity, not angular acceleration as control inputs \cite{achtelik_inversion_2013, raffler_path_2013}.
In this paper, we provide theoretic, simulation, and hardware comparisons to cascaded approaches, and present the final control law with angular acceleration as a control input.
We then extend the feedback linearization approach to incorporate a delay in the thrust control input.
Further, our derivation does not use Euler angles or quaternions and instead uses the simpler quantity of the body $z$-axis to handle the coupling between translational and rotational dynamics.

Although differential flatness and feedback linearization are closely related, differential flatness, as used in the context of feedforward linearization \cite{hagenmeyer_exact_2003}, only linearizes the system at the \textit{desired} state and corresponding input, whereas feedback linearization linearizes the system at the \textit{current} state and input. Thus, differential flatness is unable to properly linearize the system when there is large tracking error and control performance may suffer.

\citet{faessler_thrust_2017} shows that modeling the torque control input delay improves disturbance rejection, but does not consider the delay in the thrust control input and \cite{holzapfel_novel_2011} uses Pseudo Control Hedging (PCH) to mitigate delays in the control inputs, among other unmodeled effects.

Learning and feedback linearization has been explored for a long time.
As for general model learning for controls, the related works in this area can be divided into those that learn forward models and those that learn inverse models.
\citet{yesildirekt_feedback_1995} uses neural networks to learn the forward dynamics model, after which it is used in feedback linearization
and \cite{umlauft_feedback_2017} learns a forward model using a Gaussian Process (GP), whose uncertainty estimates are used to prove a convergence guarantee.
\citet{spitzer_inverting_2019} learns a forward model and uses its derivatives to compensate for the disturbance dynamics.
On the other hand, \cite{westenbroek_feedback_2020} and \cite{greeff_exploiting_2021} learn an inverse model for feedback linearization.

While model learning for quadrotors has been studied extensively, few have studied model learning for the feedback linearizing controller.
In this work, we learn a forward model for the multirotor feedback linearization controller and also consider the delay in the thrust control input.

\section{METHOD}

\subsection{System Model}

We model the multirotor as a rigid body in 3D with two control inputs: the linear acceleration along the body $z$-axis and the body angular acceleration.
To account for unmodeled dynamics and external disturbances, we include an additive acceleration component $f_e$.
\begin{align}
  \bm x = \begin{bmatrix} p \\ v \\ R \\ \omega \end{bmatrix},\quad
  \bm u = \begin{bmatrix} u \\ \alpha  \end{bmatrix},\quad
  \dot{\bm x} = \begin{bmatrix} v \\ u z + g + f_e(\xi) \\ R[\omega]_\times \\ \alpha \end{bmatrix}
  \label{eq:sys}
\end{align}
Equation \eqref{eq:sys} provides the system model. $p \in \mathbb{R}^3$ is the position to the vehicle, $v \in \mathbb{R}^3$ is the velocity, $R \in SO(3)$ is the rotation matrix that rotates vectors from the body frame into the world frame, $z = Re_3$, with $e_3 = \begin{pmatrix} 0 & 0 & 1 \end{pmatrix}^\top$ is the 3rd column of $R$, $g$ is the gravity vector, $\omega \in \mathbb{R}^3$ is the angular velocity of the body expressed in the body frame, $u \in \mathbb R$ is the acceleration along $z$, and $\alpha \in \mathbb{R}^3$ is the angular acceleration of the body expressed in the body frame.
$f_e$ is the state-dependent acceleration model correction that is learned from data as a function of an application-dependent feature vector $\xi$.

In order to account for the vehicle's rotor inertia, we add a first order exponential delay model to the vehicle thrust, $u$.
\begin{align}
  \dot u = -\tau_u (u - u^{\text{des}})
  \label{eq:thrustdelaymodel}
\end{align}
Equation \eqref{eq:thrustdelaymodel} provides a good approximation to the delay in the thrust produced by the vehicle if the relationship between force and rotor speed is linear, as rotor speeds have been experimentally shown to display first order exponential delay behavior \cite{michael_grasp_2010}. To keep the resulting feedback linearization controller simple, we do not model a delay in the angular acceleration control input.
The augmented system with the thrust control input delay component, and the thrust $u$ added to the state vector, is shown in \eqref{eq:sys2}.
\begin{align}
  \bm x = \begin{bmatrix} p \\ v \\ R \\ \omega \\ u \end{bmatrix},\quad
  \bm u = \begin{bmatrix} u^{\text{des}} \\ \alpha  \end{bmatrix},\quad
  \dot{\bm x} = \begin{bmatrix} v \\ u z + g + f_e(\xi) \\ R[\omega]_\times \\ \alpha \\ -\tau_u (u - u^{\text{des}}) \end{bmatrix}
  \label{eq:sys2}
\end{align}
Here, $u^{\text{des}}$ is the \textit{desired} linear acceleration along the body $z$-axis, and is closely tracked by the \textit{estimated} linear acceleration, $u$, according to a first order exponential delay model with time constant $\tau_u$, given by \eqref{eq:thrustdelaymodel}.

\subsection{Feedback Linearization with Dynamic Extension}

We apply feedback linearization with dynamic extension \cite{isidori_nonlinear_1995} to the above multirotor system model \eqref{eq:sys2}, with the objective of controlling the system position $p$.
The relative degree of the multirotor system \eqref{eq:sys2} is not defined as written, since $u^{\text{des}}$ first appears in the third derivative of the vehicle position, while $\alpha$ first appears in the fourth derivative, through $\ddot z$.

Dynamic extension is a technique to transform a system without a relative degree into one with a relative degree through the use of auxiliary state variables that delay the introduction of control inputs. In the traditional multirotor case, we need to delay the body thrust $u$ twice, so that it appears in concert with the angular acceleration \cite{mistler_exact_2001}. For the system model we consider, which models thrust delay, we only need to delay the linear acceleration input once. For that, we move $u^{\text{des}}$ into the state and replace it with its derivative, $\dot{u}^{\text{des}}$.

\newcommand{\cu}{\dot{u}^{\text{des}}}

Now to solve for the control inputs $\cu$ and $\alpha$, we differentiate the output four times.
\begin{align}
  \dot p &= v \\
  \ddot p \equiv a &= u z + g + f_e(\xi) \label{eq:acc} \\
  p^{(3)} \equiv j &= \dot u z + u \dot z + \dot f_e(\xi) \label{eq:jerk} \\
  p^{(4)} \equiv s &= \ddot u z + 2 \dot u \dot z + u \ddot z + \ddot f_e(\xi)
  \label{eq:snap}
\end{align}

\subsubsection[Computing the thrust control input]{Computing the thrust control input $u^{\text{des}}$}

We can project the snap to the body $z$-axis to solve for $\ddot u$, noting that $z^\top \dot z = 0$ and $z^\top \ddot z = -\dot z ^\top \dot z$.
\begin{align}
  \ddot u &= s^\top z + u \dot z ^\top \dot z - \ddot f_e(\xi)^\top z \nonumber \\
          &= \left(s - \ddot f_e(\xi)\right)^\top z + u \dot z ^\top \dot z
  \label{eq:uddot}
\end{align}
We also note that, according to the thrust delay model \eqref{eq:thrustdelaymodel},
\begin{align}
  \ddot u = -\tau_u (\dot u - \cu)\text{.}
  \label{eq:dthrustdelaymodel}
\end{align}
Combining \eqref{eq:uddot}, \eqref{eq:dthrustdelaymodel}, and \eqref{eq:thrustdelaymodel}, we have
\begin{align}
  \cu = \frac{(s - \ddot f_e(\xi) )^\top z + u \dot z ^\top \dot z}{\tau_c} - \tau_c(u - u^{\text{des}})
  \label{eq:udotdes}
\end{align}
The thrust control input $u^{\text{des}}$ and the estimated thrust $u$ are computed from \eqref{eq:udotdes} and \eqref{eq:thrustdelaymodel} respectively, using numerical integration.

\subsubsection[Computing the angular acceleration control input]{Computing the angular acceleration control input $\alpha$}
\newcommand{\wom}{{\omega_{\mathcal{W}}}}
\newcommand{\waa}{{\alpha_{\mathcal{W}}}}

From \eqref{eq:snap}, we can solve for $\ddot z$.
\begin{align}
  \ddot z = \frac{1}{u}\left( s - \ddot f_e(\xi) - \ddot u z - 2 \dot u \dot z \right)
\end{align}
Noting that $\dot z = \wom \times z$
and $\ddot z = \waa \times z + \wom \times \dot z$,
with $\wom = R\omega$ and $\waa = R\alpha$ the angular velocity and angular acceleration of the vehicle in the world frame,
the vector triple product gives
\begin{align}
  z \times \ddot z
    &= z \times (\waa \times z) + z \times (\wom \times (\wom \times z)) \nonumber \\
    &= \waa - (\alpha_{\mathcal{W}} ^\top z) z + (\omega_{\mathcal{W}} ^\top z) z \times \wom .
\end{align}
Thus the angular acceleration less that along the $z$ axis, or the angular acceleration along the body $x$ and body $y$ axes, denoted $\alpha^{xy}$, can be found using
\begin{align}
  \alpha^{xy}_{\mathcal{W}} &= z \times \ddot z - (\omega_{\mathcal{W}} ^\top z) z \times \omega \nonumber \\
  &= \frac{1}{u}\left(z \times (s - \ddot f_e(\xi)) - 2\dot u (z \times \dot z) \right) + (\omega_{\mathcal{W}} ^\top z) \wom \times z \nonumber \\
  &= \frac{1}{u}\left(z \times (s - \ddot f_e(\xi)) - 2\dot u \omega^{xy}_{\mathcal{W}} \right) + (\omega_{\mathcal{W}} ^\top z) \wom \times z
  \label{eq:accxy}
\end{align}
Here, $\omega^{xy}_{\mathcal{W}}$ is the angular velocity less that along the $z$ axis.
The component of the angular acceleration along the body $z$-axis is undetermined by the position and its derivatives and thus can be chosen to follow a desired yaw trajectory.
We omit the derivation for brevity.

\subsubsection{Linear Feedback Controller}

Eqs. \eqref{eq:udotdes} and \eqref{eq:accxy} together provide the control inputs that are needed to achieve snap $s$.
Snap is typically the result of a linear feedback controller that tracks a desired position, velocity, acceleration, jerk, and snap, as shown in \eqref{eq:snapcontrol}, where $x^{\text{err}} = x - x^{\text{des}}$.
\begin{align}
  s &= -k_1p^{\text{err}} - k_2v^{\text{err}} - k_3a^{\text{err}} - k_4j^{\text{err}} + s^{\text{des}}
  \label{eq:snapcontrol}
\end{align}
Gains for this controller, $k_i \in \mathbb{R}^{3 \times 3}$, can be chosen by any linear control technique, such as LQR and pole placement.
When the acceleration and jerk are not readily available, as is the case with most state estimators, Eqs. \eqref{eq:acc} and \eqref{eq:jerk} can be used for feedback.
Further, to support implementation on an embedded platform, the disturbance model $f_e$ and its derivatives can be evaluated on a ground control station.
In that case, let $s^{\text{ff}} = -k_1p^{\text{err}} - k_2v^{\text{err}} + k_3(a^{\text{des}} - f_e(\xi)) + k_4(j^{\text{des}} - \dot f_e(\xi)) + s^{\text{des}} - \ddot f_e(\xi)$ and the snap becomes
\begin{align}
  s &= s^{\text{ff}} - k_3(uz + g) - k_4(\dot u z + u \dot z) + \ddot f_e(\xi)
  \label{eq:swff}
\end{align}
We can combine \eqref{eq:swff} with \eqref{eq:accxy} to further simplify the expression for angular acceleration along the body $x$ and $y$ axes.
{
  \small
\begin{align}
  \alpha^{xy}_{\mathcal{W}} = \frac1u \left(z \times s^{\text{ff}} - k_3 z \times g - 2\dot u \omega^{xy}_{\mathcal{W}}\right) - k_4 \omega^{xy}_{\mathcal{W}} - (\omega_{\mathcal{W}}^\top z) z \times \omega_{\mathcal{W}}
  \label{eq:angxyfull}
\end{align}
}
To obtain the control input $\alpha_{xy}$ in the body frame, we left-multiply \eqref{eq:angxyfull} by $R^\top$ to obtain.
{ \small
\begin{align}
  \alpha^{xy} = \frac{1}{u} \left(
  e_3 \times s^{\text{ff}}_{\mathcal{B}} - k_3 e_3 \times g_{\mathcal{B}} -2\dot u \omega^{xy} \right) - k_4 \omega^{xy} - \omega^z e_3 \times \omega
  \label{eq:angxybody}
\end{align} }
We similarly transform the computation of $\dot u_{\text{des}}$ into the body frame, after combining \eqref{eq:swff} with \eqref{eq:udotdes}, for a simpler implementation on an embedded platform.
{\small
  \begin{align}
    \cu = \frac{s^{\text{ff}\top}_{\mathcal{B}} e_3 - k_3( u + g_{\mathcal{B}}^\top e_3) - k_4 \dot u + u||\omega^{xy}||^2}{\tau_c} - \tau_c(u - u^{\text{des}})
  \label{eq:udotdesbody}
\end{align}}
Note that implementing \eqref{eq:angxybody} and \eqref{eq:udotdesbody} only requires knowing the gravity vector, the angular velocity, and $s^{\text{ff}}$ in the body frame.
The gravity vector in the body frame is directly measured by an IMU that is at the center of the body frame and the angular velocity in the body frame is directly measured by a gyroscope that is anywhere on the rigid body.
Further, this controller requires no trigonometric operations, such as $\sin$, $\cos$, $\tan$, or their inverses, and can run very quickly on an embedded system.
Notably, the control law does not depend on the yaw trajectory of the vehicle, although the full orientation is required to compute $s^{\text{ff}}$ in the body frame if the position feedback and trajectory planning are with respect to the fixed frame.

\subsection{Acceleration Model Learning}

To estimate $f_e(\xi)$ from vehicle trajectory data, we fit a model to differences between the observed and the predicted acceleration at every time step.
We use a subset of the state as input to the model: the position $p$ and velocity $v$.
Although this limits the types of disturbances that can be accounted to those that are functions of vehicle position and velocity, and notably not disturbances that are functions of the control input, this is a suitable choice to highlight the proposed control algorithm, given the experiments performed in Section \ref{sec:results}.
The observed acceleration is computed using finite-differences of the estimated vehicle velocity sampled in intervals of $\Delta T$, while the predicted acceleration is $uz + g$.
Thus, the training examples used to fit the model are pairs of data points $(\xi_t, y_t)$, defined below.
\begin{align}
  \xi_t &= \begin{pmatrix} p_t & v_t \end{pmatrix}^\top \\
  y_t &= \frac{1}{\Delta T}(v_t - v_{t - 1}) - (u_{t - 1}z_{t - 1} + g)
\end{align}

In this work, we use Incremental Sparse Spectrum Gaussian Process Regression (ISSGPR) \cite{gijsberts_real-time_2013} as the regression strategy.
ISSGPR projects the input data into a nonlinear feature space defined by sinusoids with random frequencies \cite{rahimi_random_2007} and then applies regularized linear regression incrementally using rank-one matrix updates.
Let $d_x$ be the dimension of the input vector $\xi$ and let $N$ be the number of random frequencies chosen.
The random frequencies $\Omega \in \mathbb{R}^{N \times d_x}$ can be generated by multiplying a $N$ by $d_x$ matrix of univariate Gaussians by $\text{diag}(M)$, where each entry in $M \in \mathbb{R}^{d_x}$ is the inverse of the characteristic length scale of the corresponding input dimension.
The length scales allow for the adjusting of the relative importance of each input dimension and are thus hyperparameters that need to be adapted to the application.

The output of the model is shown below. Here, $W~\in~R^{2N \times 3}$ is the matrix of weights learned by linear regression, where each column corresponds to a dimension in the output.
\begin{align}
  f_e(\xi) = \frac{1}{\sqrt{N}}W^\top \begin{pmatrix} \cos(\Omega \xi) \\ \sin(\Omega \xi) \end{pmatrix}
\end{align}

The derivatives of $f_e$ with respect to the input $\xi$, which are required to implement the proposed control strategy, are shown below.
Here, $\xi_k$ is the $k$'th entry of $\xi$, $\Omega_k$ is the $k$'th column of $\Omega$, and $\odot$ is the Hadamard, or component-wise, product.
\begin{align}
  \frac{\partial f_e(\xi)}{\partial \xi_i} = \frac{1}{\sqrt{N}} W^\top \begin{pmatrix*}[r]
           -\sin(\Omega \xi) \odot \Omega_i \\
            \cos(\Omega \xi) \odot \Omega_i \end{pmatrix*}
\label{eq:modd1}
\end{align}
\begin{align}
  \frac{\partial^2 f_e(\xi)}{\partial \xi_i\xi_j} = \frac{1}{\sqrt{N}} W^\top \begin{pmatrix*}[r]
           -\cos(\Omega \xi) \odot \Omega_i \odot \Omega_j \\
           -\sin(\Omega \xi) \odot \Omega_i \odot \Omega_j \end{pmatrix*}
\label{eq:modd2}
\end{align}
Using \eqref{eq:modd1} and \eqref{eq:modd2}, $\dot f_e$ and $\ddot f_e$ can be computed using the chain rule and knowledge of $\dot \xi$ and $\ddot \xi$, as depicted in Fig.~\ref{fig:pictoral}.

\section{EXPERIMENTS}
\label{sec:results}

We aim to show the following three results:
\begin{itemize}
  \item [\textbf{R1}] The feedback linearization controller presented handles large step responses with less deviations along unexcited axes than traditional cascaded approaches.
  \item [\textbf{R2}] Accounting for rotor delay using a delayed thrust model increases performance of the feedback linearization controller during step inputs in the presence of control input delays.
  \item [\textbf{R3}] The learned acceleration error model improves control performance of the feedback linearization controller in the presence of unmodeled dynamics and external disturbances.
\end{itemize}

\subsection{Position and Yaw Step Response}
\label{sec:results-yawstep}

\begin{figure}
  \begin{center}
    \includegraphics[width=0.48\textwidth]{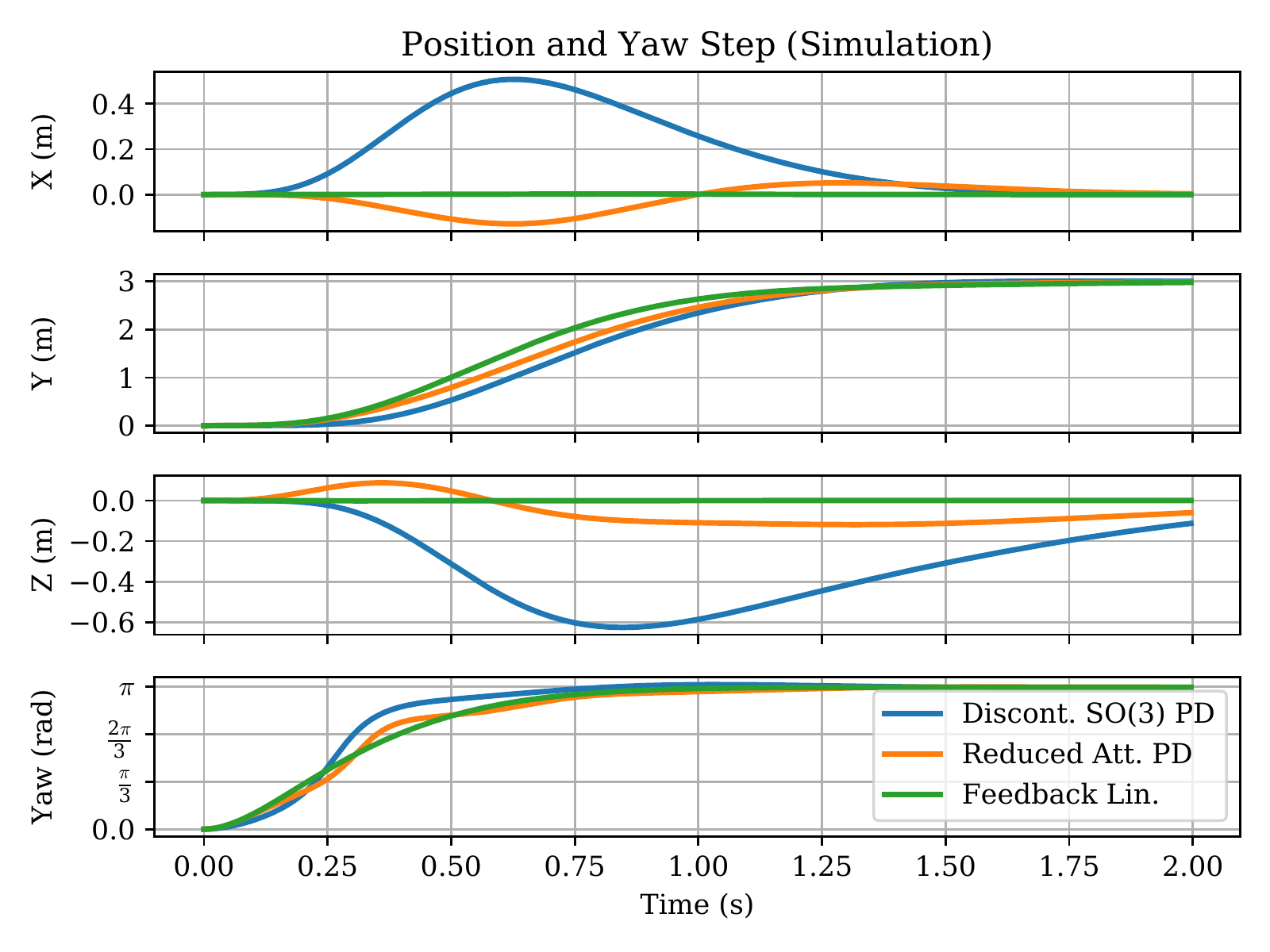}
  \end{center}
  \caption{Position and yaw during a step from $(x, y, z, \psi) = (0, 0, 0, 0)$ to $(0, 3, 0, \pi - 0.01)$ for a traditional cascaded controller (blue), reduced attitude control (orange), and feedback linearization (green). Only feedback linearization follows the straight path from $p = (0, 0, 0)$ to $(0, 3, 0)$.}
  \label{fig:simyawstep}
\end{figure}

To highlight the advantages of feedback linearization over traditional controllers during large tracking error, we simulate the vehicle executing a simultaneous step along position and yaw.
We compare against traditional discontinuous SO(3) control \cite{fresk_full_2013}, as well as reduced attitude control \cite{brescianini_tilt-prioritized_2020}.
The simulation performance of feedback linearization and the baseline controllers for a step response from
$(x, y, z, \psi) = (0, 0, 0, 0)$ to $(0, 3, 0, \pi - 0.01)$
is shown in Fig.~\ref{fig:simyawstep}.
The traditional attitude controller, which uses a metric on SO(3), suffers deviations along the $x$ and $z$-axes, despite the position step lying along the $y$-axes.
Although reduced attitude control is able to reduce these deviations, feedback linearization eliminates them entirely, thus showing \textbf{R1}.

\subsection{Hardware Setup}

The quadrotor used in the hardware experiments weighs \SI{650}{\gram} and has a thrust to weight ratio of approximately 5.
Position feedback is provided at \SI{100}{\hertz} and sent to the vehicle via a \SI{2.4}{\giga\hertz} radio, while attitude control is run onboard the vehicle at \SI{500}{\hertz}.

Feedback linearization gains for the vehicle are $k_1 = \text{diag}(1040, 1040, 1900)$, $k_2 = \text{diag}(600, 600, 1140)$, $k_3 = 190I$, and $k_4 = 25I$. These were chosen to provide the same first order response in angular acceleration as the cascaded gains $k_p = \text{diag}(5.47, 5.47, 10.0)$, $k_v = \text{diag}(3.16, 3.16, 6.0)$, $k_\theta = \text{diag}(190, 190, 30)$, and $k_\omega = \text{diag}(25, 25, 10)$. Note that the first order response in thrust cannot be matched as it is a dynamic first order process for the feedback linearization controller.

\subsection{Control Input Delay}

To test the performance of the control input delay mitigation strategy on the hardware vehicle, we execute ten \SI{3}{\meter} steps in position with the feedback linearization controller and various values of $\tau_u$.
\begin{figure}
  \begin{center}
    \includegraphics[width=9cm]{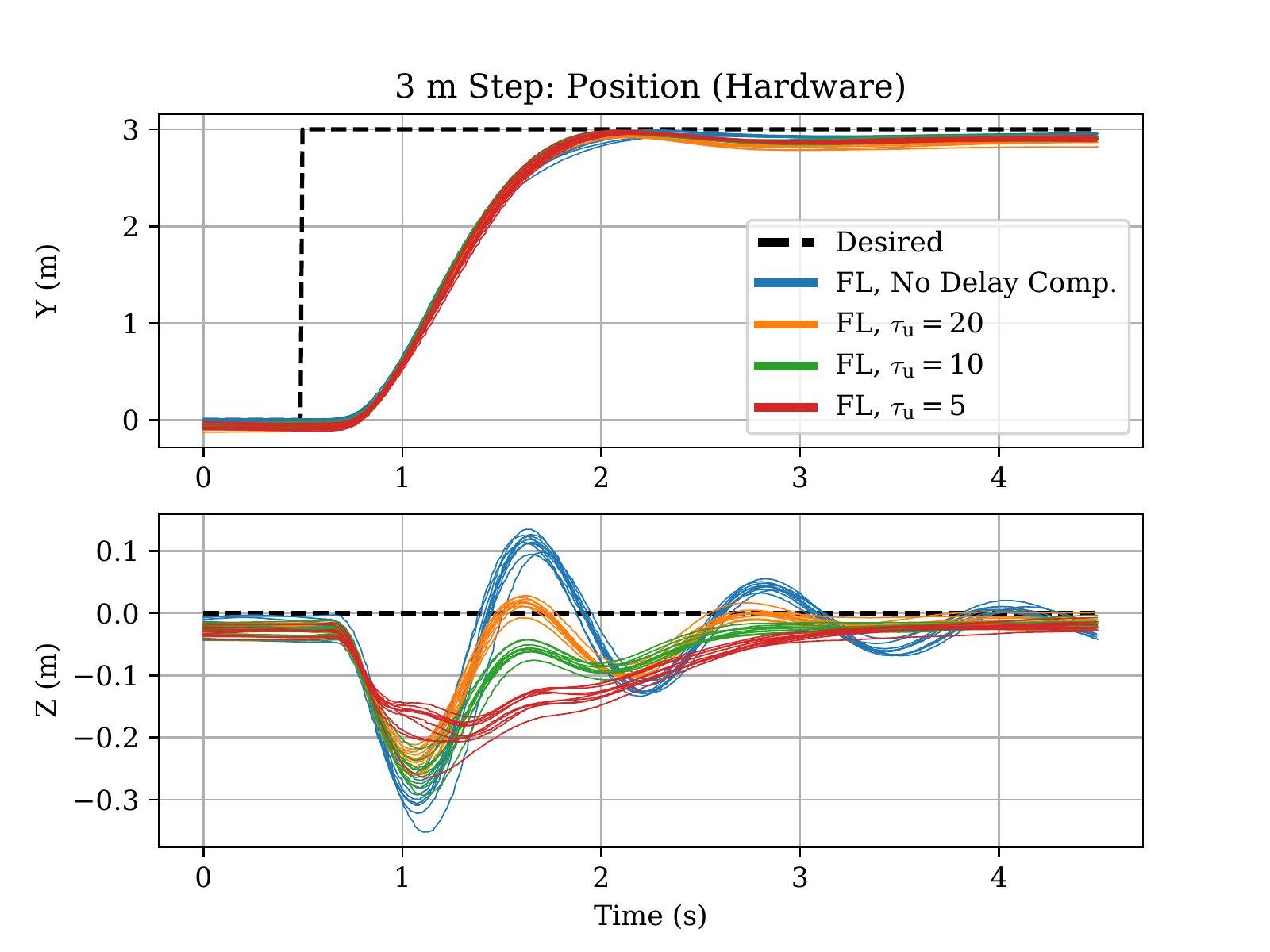}
  \end{center}
  \caption{Position during ten \SI{3}{\meter} sideways steps on the hardware platform for various feedback linearizaton configurations. Our proposed approach, shown here assuming a thrust delay with time constants of $5$, $10$, and $20$, maintains height better, and reduces oscillations compared to standard feedback linearization.}
  \label{fig:sidestep_side_hardware}
\end{figure}
Figure \ref{fig:sidestep_side_hardware} shows the trajectory side view for the \SI{3}{\meter} steps executed on the hardware platform.
Ten trials are executed for the following: feedback linearization without delay compensation and feedback linearization assuming thrust control input delays of $\tau = 5$, $10$, and $20$.
We see that considering the thrust input delay significantly reduces the altitude oscillations and altitude error when compared to standard feedback linearization, supporting \textbf{R2} on the hardware platform.
For the remaining hardware experiments, we use $\tau_u = 10$, as it provides good reduction in oscillations, without greatly overestimating the delay.

\subsection{Learned Acceleration Error Model}
\begin{figure}
  \begin{center}
    \includegraphics[trim=0 20 0 75, clip, width=8.09cm]{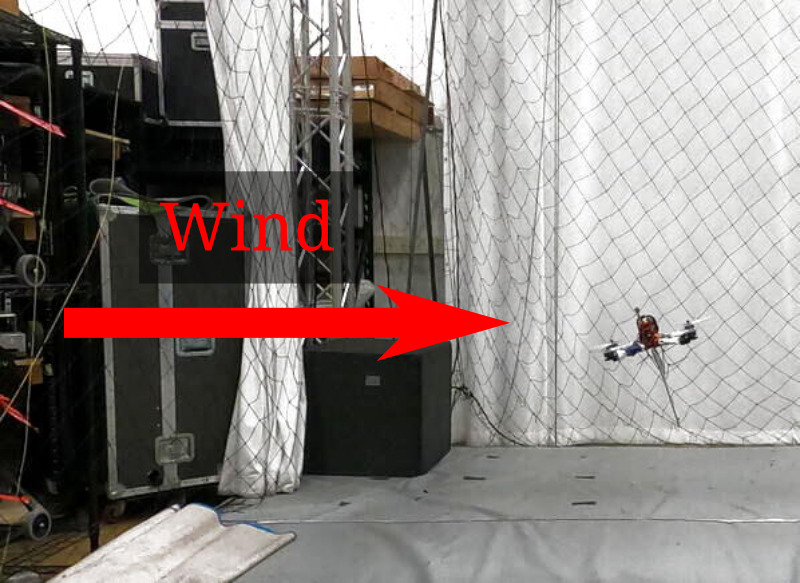}
    \includegraphics[trim=10 0 0 0, clip, width=4cm]{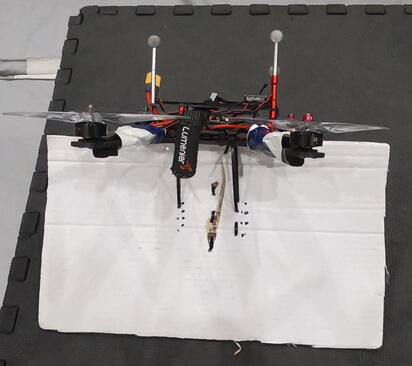}
    \includegraphics[trim=0 0 0 10, clip, width=4cm]{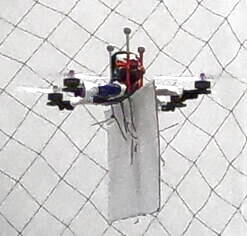}
  \end{center}
  \caption{
    The quadrotor vehicle during the yaw in place experiment, with wind from two industrial fans coming from the left (top), the cardboard plate attached to amplify wind effect (bottom left), and the vehicle hovering with the cardboard plate without wind (bottom right).
  }
  \label{fig:vehicle}
\end{figure}
We test the effectiveness of the learned acceleration error model in the presence of unmodeled dynamics and external disturbances by attaching a cardboard plate to the quadrotor while flying through a turbulent wind field, as shown in Fig.~\ref{fig:vehicle}.
The cardboard plate serves to accentuate disturbances due to the wind field.

We test the approach in two scenarios:
\begin{enumerate}
  \item 3D weave pattern through the wind field
  \item Fast yaw in place in the wind field
\end{enumerate}
\subsubsection{3D Weave Through Wind Field}
\begin{figure}
  \begin{center}
    \includegraphics[trim=0 0 0 25, clip, width=7cm]{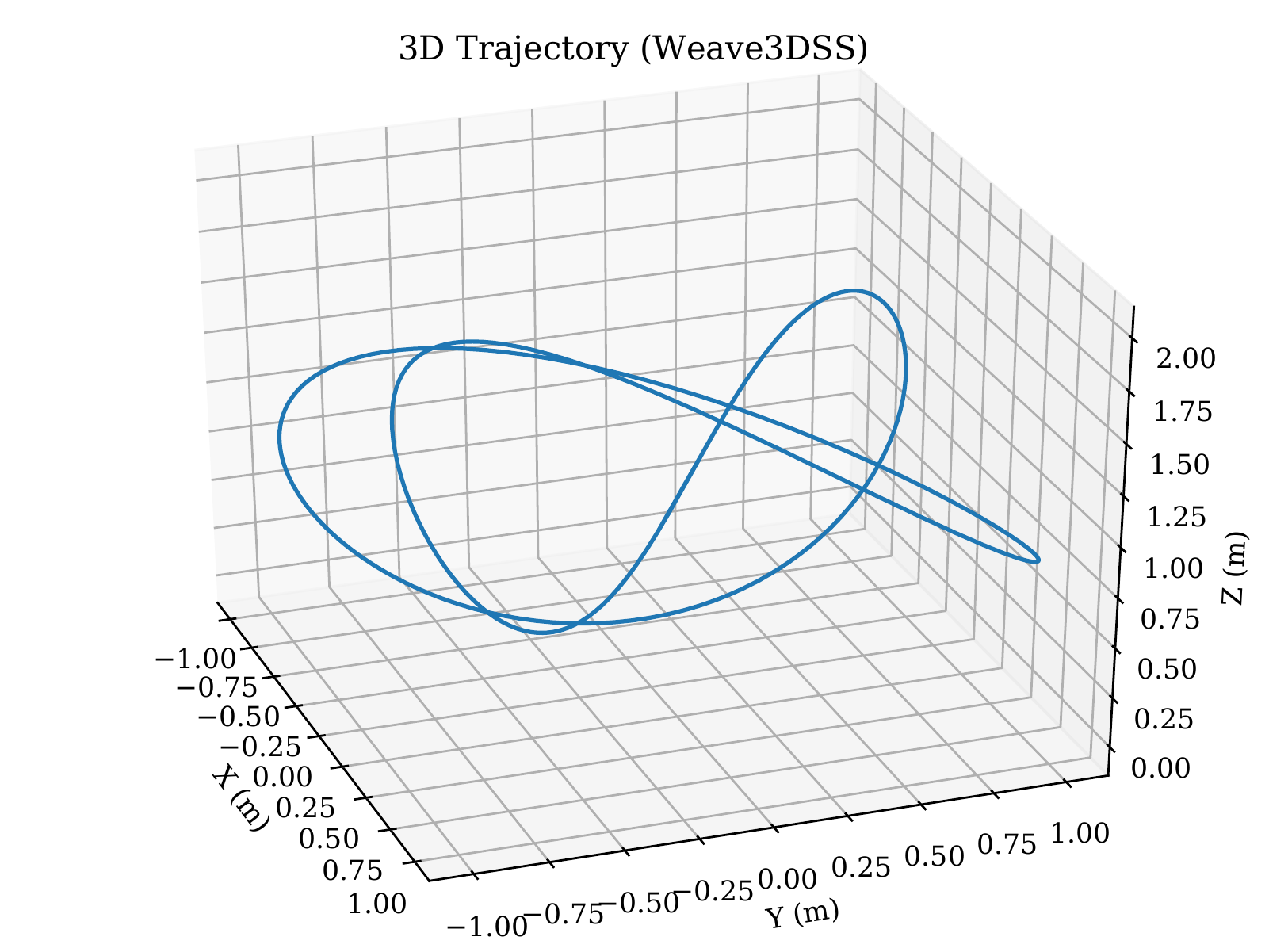}
  \end{center}
  \caption{The 3D weave trajectory used in the experimental evaluation of the control strategy.}
  \label{fig:weave}
\end{figure}
The weave trajectory, shown in Fig.~\ref{fig:weave}, has a maximum velocity of \SI{2.7}{\meter\per\s}, a maximum acceleration of \SI{5.5}{\meter\per\s\squared}, and moves in a sinusoidal pattern along $x$, $y$, and $z$.
For this experiment, the performance of the proposed model learning strategy is compared to a controller using \textit{L1 Adaptive Control} (L1AC) \cite{michini_l1_2009} for acceleration disturbance compensation and a cascaded PD controller running the model learning strategy from \cite{spitzer_inverting_2019}.
Since we are correcting for model errors during trajectory following, a low tracking error will mean that differences in feedback strategy are negligible, and we expect the cascaded PD approach to perform similarly to the feedback linearization approach.
For this experiment, we use $N = 50$ random frequencies for ISSGPR and length scales of $1.0$ for each dimension of the position and velocity.
\begin{figure}
  \begin{center}
    \includegraphics[width=9cm]{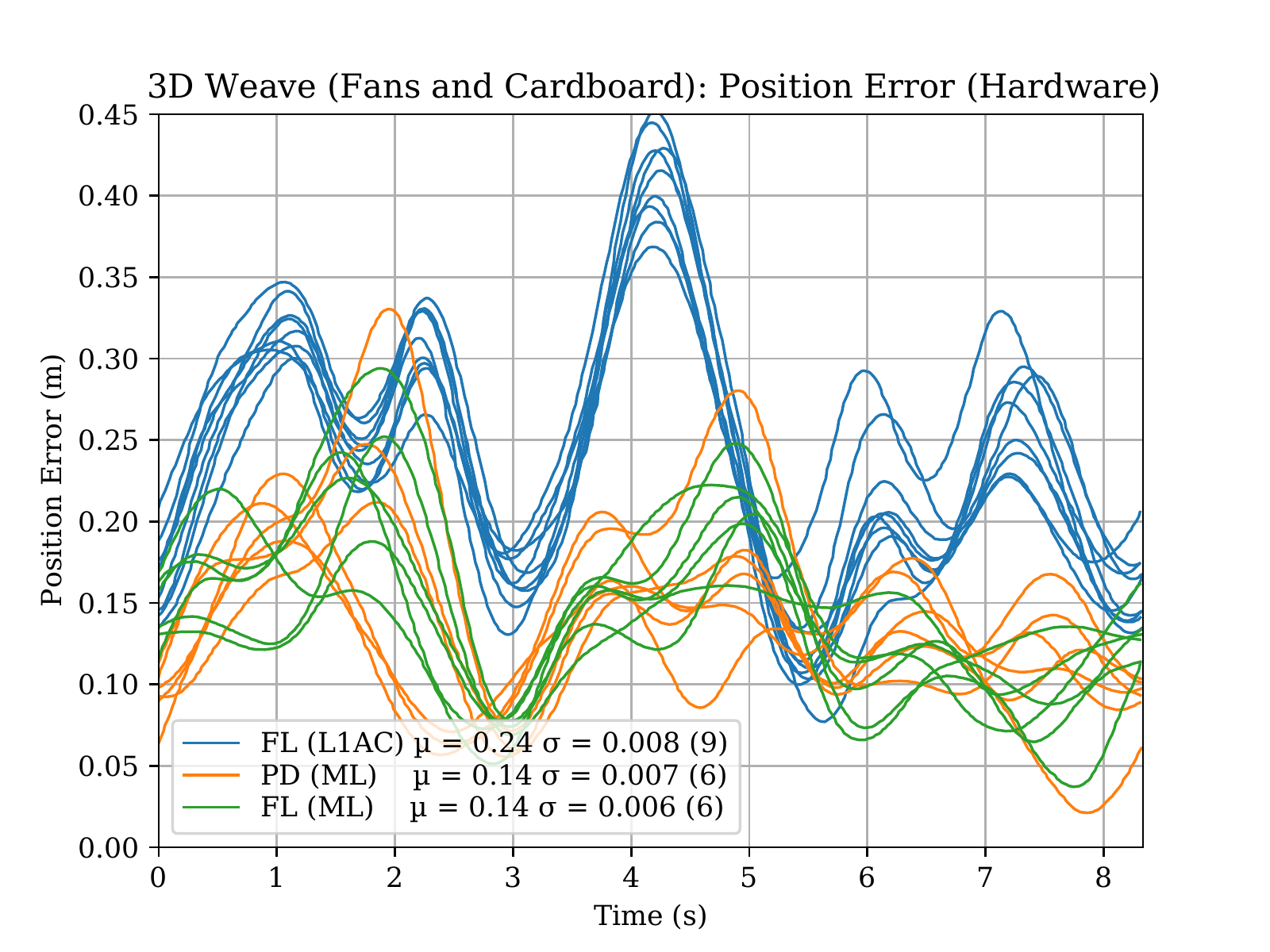}
  \end{center}
  \caption{Smoothed position error over multiple trials of the 3D weave for the baseline adaptive controller (blue), the proposed model learning strategy with PD control (orange), and the proposed model learning strategy with feedback linearization (green). The model learning strategy reduces the tracking error by roughly \SI{40}{\percent} compared to the adaptive controller. As expected, both cascaded control and feedback linearization perform well.}
  \label{fig:hwweave}
\end{figure}

The position error during the weave trajectories for each strategy is shown in Fig.~\ref{fig:hwweave}.
Only the second and third trials (each trial consists of three circuits of the weave) for the two strategies using model learning are included to highlight the performance after the acceleration error model has converged.
The model learning strategy is able to reduce the error by around \SI{40}{\percent} relative to the adaptive controller and as expected, feedback linearization performs similarly to the cascaded approach.
\begin{figure}
  \begin{center}
    \includegraphics[width=8cm]{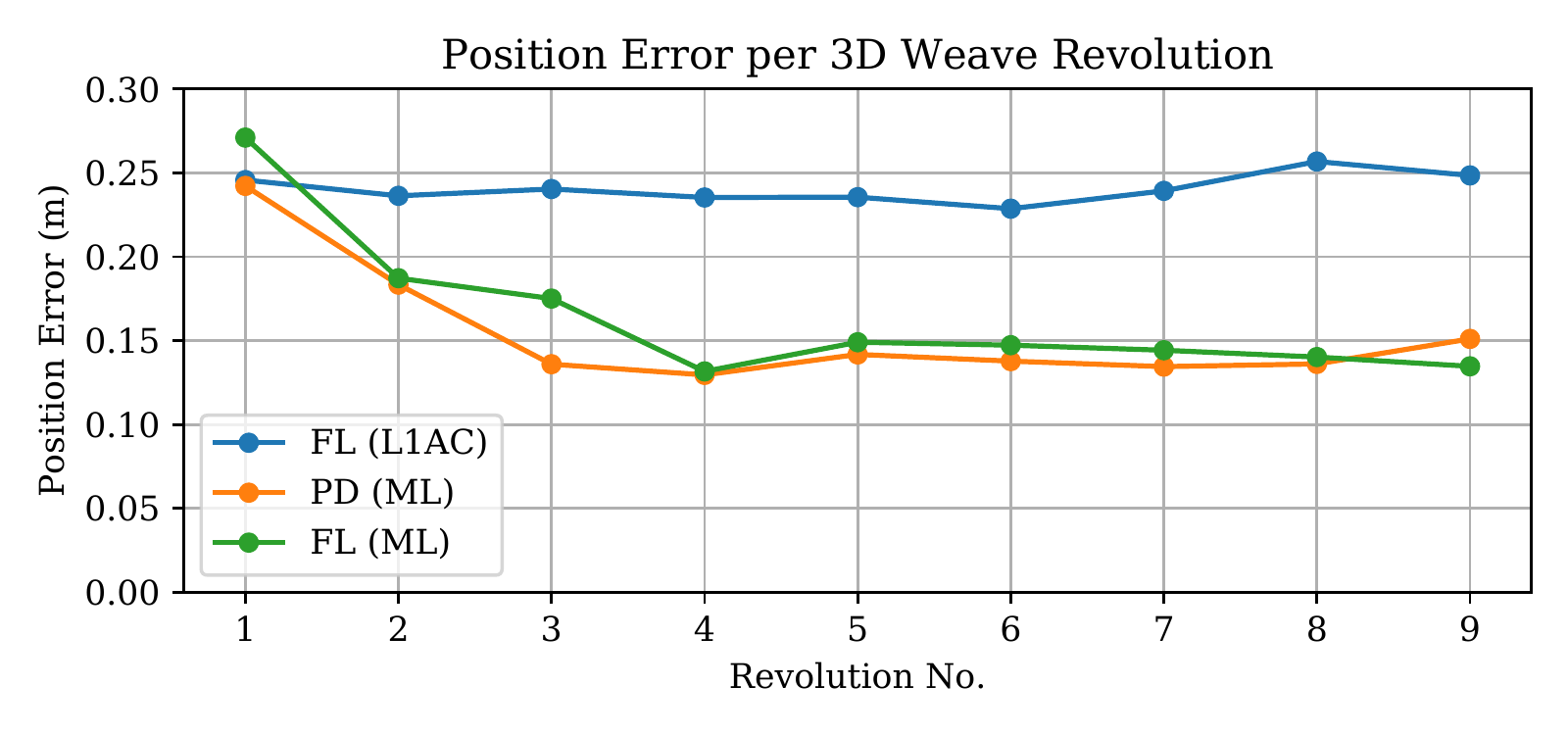}
  \end{center}
  \caption{
    Position per revolution of the 3D weave trajectory for the baseline adaptive controller (blue), the proposed model learning strategy with PD control (orange), and the proposed model learning strategy with feedback linearization (green). The learned model converges quickly, stabilizing the error to the minimum attained after two revolutions around the weave.
}
  \label{fig:hwmlrev}
\end{figure}

Figure~\ref{fig:hwmlrev} shows the average position error for each circuit of the weave for the three strategies.
After two circuits, the learned model has converged and subsequent circuits have the minimum error seen throughout the experiment.
Thus, the model learning strategy converges quickly without sacrificing stability as it is learning.

\subsubsection{Yaw in Place in Wind Field}
For the second test, seen in Fig.~\ref{fig:vehicle}, the vehicle yaws in place in the wind field at a rate of \SI{120}{\degree\per\s} for 4 revolutions.
To enable the vehicle to model the changing acceleration disturbance as the vehicle yaws, we include two extra features in $\xi$ in addition to the position and velocity: $\sin(\psi)$ and $\cos(\psi)$.
The length scales used for this experiment are 5.0 for position and velocity and 0.5 for the yaw terms.

In this test, we compare the performance of the proposed model learning strategy with \textit{L1 Adaptive Control} as before, and as an ablation test, with the proposed strategy but with disturbance dynamics neglected, i.e. $\dot f_e = \ddot f_e = 0$.
\begin{figure}
  \begin{center}
    \includegraphics[width=9cm]{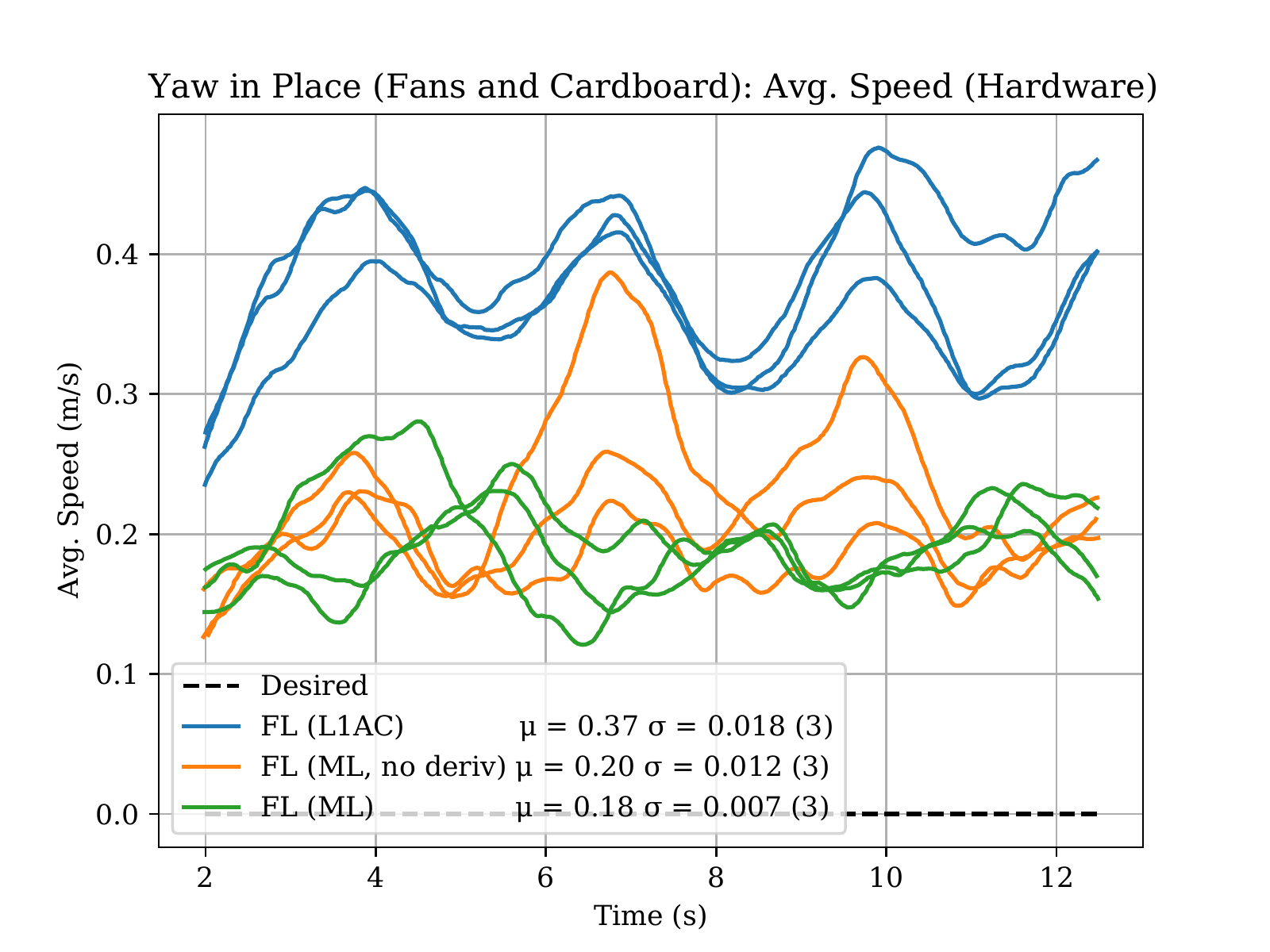}
  \end{center}
  \caption{Average speed over multiple trials of the yaw in place test at \SI{120}{\degree\per\s} for the baseline adaptive controller (blue), the proposed model learning strategy without disturbance dynamics (orange), and the proposed model learning strategy with full disturbance dynamics compensation (green). The model learning strategy is able to stay still with roughly \SI{50}{\percent} lower speed compared to the adaptive controller. Including disturbance dynamics further helps the vehicle maintain position.}
  \label{fig:hwyaw}
\end{figure}

The average speed during the experiment for all three strategies is shown in Fig.~\ref{fig:hwyaw}.
As before, the first trajectory for the model learning strategies is omitted, to remove transients from the model learning process.
The adaptive controller struggles to stay still since it cannot react to the disturbance fast enough.
The model learning strategies can predict the disturbance ahead of time, and reduce the speed by roughly \SI{50}{\percent}, showing \textbf{R3}.
Compensating for disturbance dynamics helps to further reduce the speed error.

\begin{figure}
  \begin{center}
    \includegraphics[width=8cm]{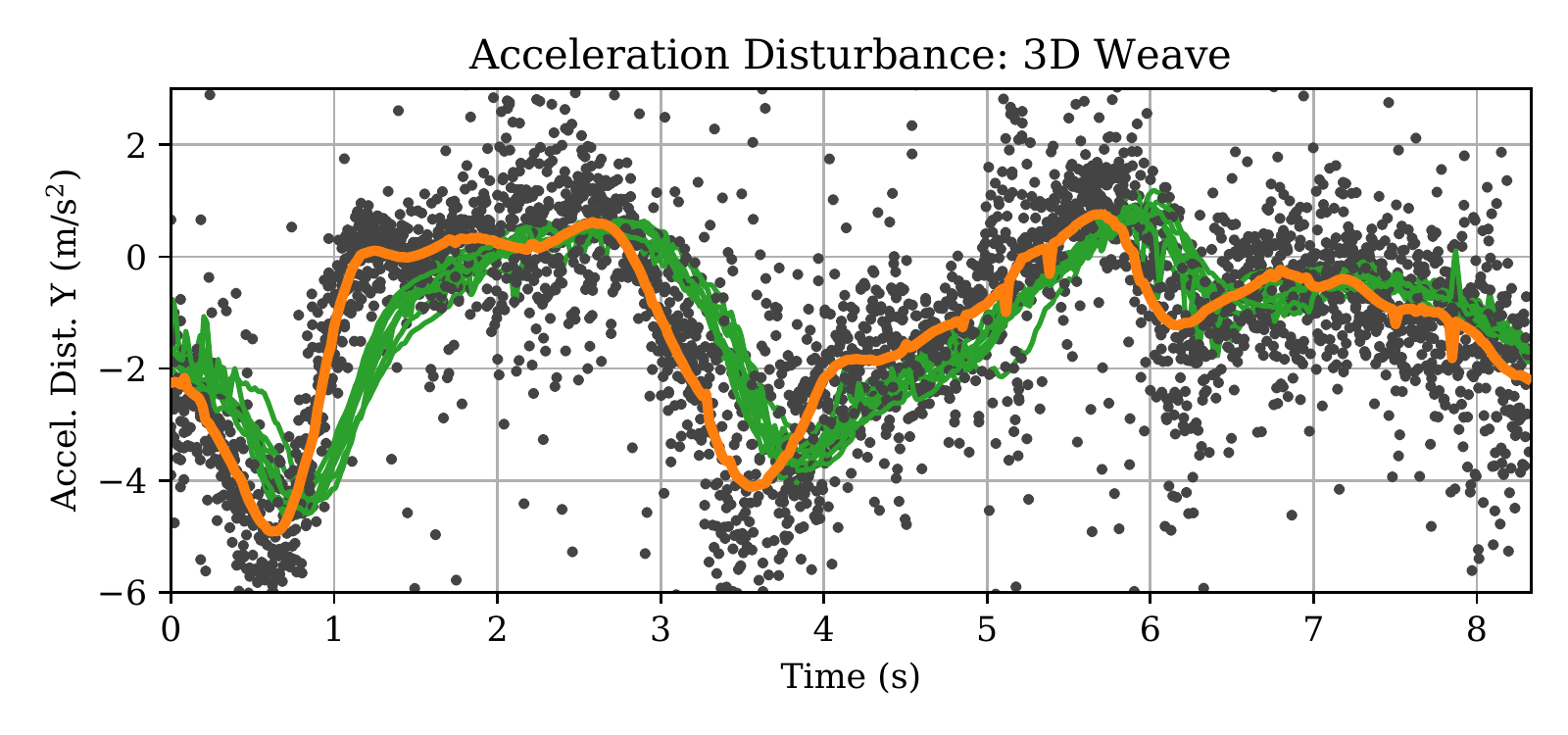} \\
    \vspace{-0.5em}
    \includegraphics[width=8cm]{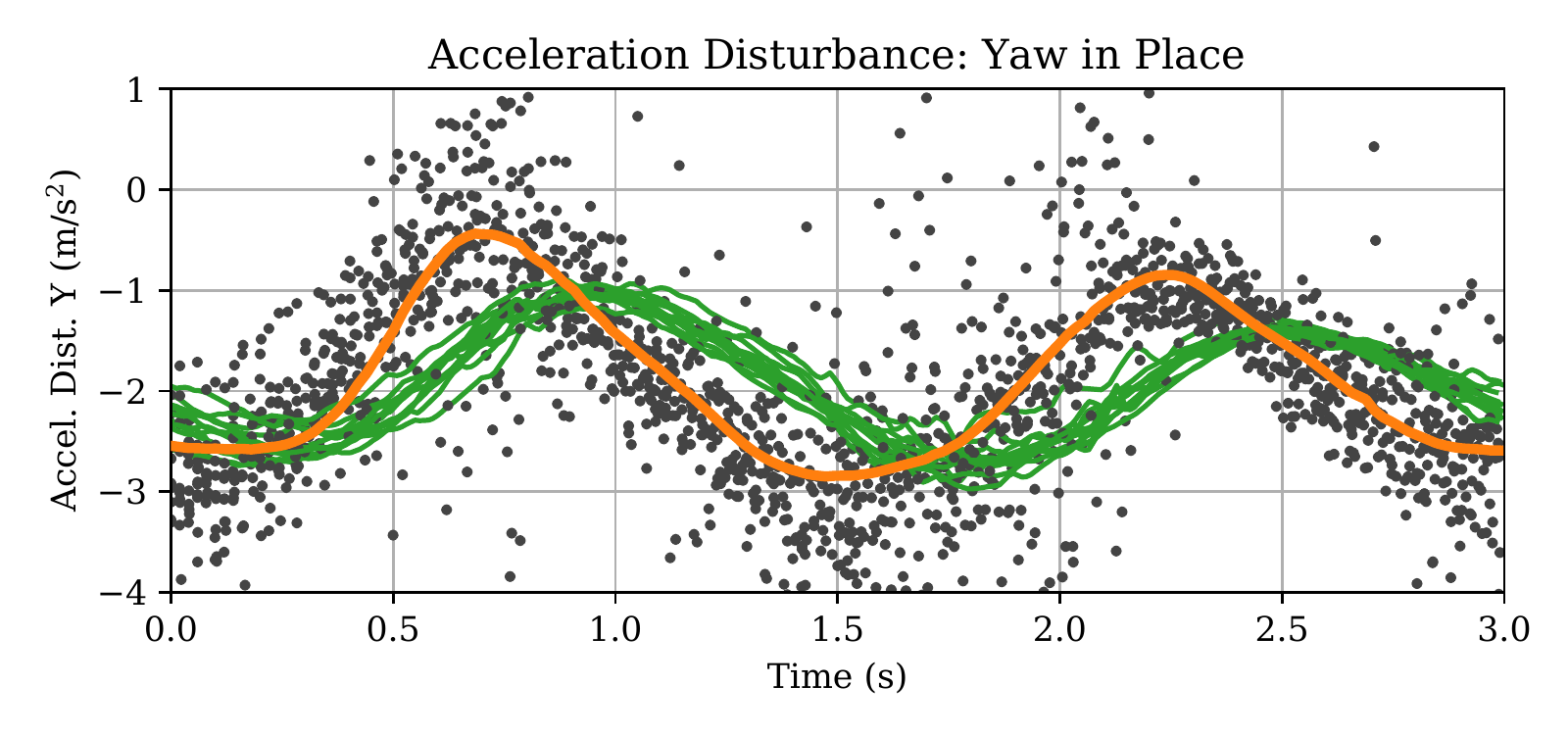} \\
    \vspace{-0.5em}
    \includegraphics[width=5cm]{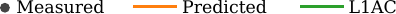}
  \end{center}
  \caption{
    The raw acceleration disturbance measurements (black), the disturbance estimated by the adaptive controller (green), and the disturbance predicted by our learned model (orange) for the weave (top) and yaw in place (bottom) experiments. The disturbance predicted by the learned model captures the mean of the disturbance well, while the disturbance estimated by the adaptive controller lags.
}
  \label{fig:hwml}
\end{figure}

Figure~\ref{fig:hwml} shows the raw acceleration disturbance measurements, the disturbance estimated by the adaptive controller, and the disturbance predicted by our learned model for the two hardware experiments.
The adaptive controller's disturbance estimate lags the true disturbance, while the learned model's prediction accurately captures the mean of the disturbance, leading to superior control performance.
Increasing the adaptation rate (bandwidth) of the baseline disturbance estimator to attempt to reduce the lag would introduce noise and instability into the system.
The learning algorithm appropriately averages sensor data to learn a smooth model from noisy input points.

\section{CONCLUSION}

We have presented a feedback linearization controller for the multirotor that outperforms existing cascaded approaches during aggressive step responses in simulation (\textbf{R1}), mitigates the negative impact of control input delays (\textbf{R2}), and reduces errors arising from external disturbances through modeling and inversion of the learned disturbance (\textbf{R3}).

Limitations arise from the need to model the disturbance as a function of the state space.
The current incremental strategy cannot properly handle time-varying disturbances that often arise in practical conditions.
Recent advances in deep learning can enable more capable model learning methods that require less hand-engineered features and parameters.

Future work should consider the delay in the angular acceleration control input, as well as the delay in the thrust control input, which is considered here.
Further, a theoretical analysis of the control performance differences between feedforward linearization and feedback linearization in the presence of model error will enable intelligent selection of controllers and possible incorporation of uncertainty bounds into the control strategy.


{
\bibliographystyle{IEEEtranN_titleURL}
\bibliography{IEEEabrv,refs}
}

\end{document}